\RequirePackage{fix-cm}
\PassOptionsToPackage{sort&compress}{natbib}
\documentclass[twocolumn,natbib]{svjour3}
\usepackage{stackengine}

\usepackage{amsmath}
\usepackage{bm}

\smartqed  
\usepackage{graphicx}
\usepackage[caption=false,labelformat=simple]{subfig}

\usepackage{latexsym}
\usepackage[export]{adjustbox}
\usepackage[strings]{underscore}
\setcitestyle{square,comma,numbers}
\journalname{Progress in Artificial Intelligence}
\begin{document}

\title{Improving Transparency of Deep Neural Inference Process}


\author{Hiroshi Kuwajima \and Masayuki Tanaka \and Masatoshi Okutomi}


\institute{H. Kuwajima \at
              DENSO CORPORATION \\
              Tokyo Institute of Technology \\
              \email{hiroshi\_kuwajima@denso.co.jp} 
           \and
           M. Tanaka \at
              National Inst. of Advanced Industrial Science and Technology \\
              \email{masayuki.tanaka@aist.go.jp}
           \and
           M. Okutomi \at
              Tokyo Institute of Technology \\
              \email{mxo@sc.e.titech.ac.jp}
}

\date{Received: 06 Sep 2018 / Accepted: 26 Feb 2019}

\maketitle

\begin{abstract}
Deep learning techniques are rapidly advan-ced recently, and becoming a necessity component for widespread systems. 
However, the inference process of deep learning is black-box, and not very suitable to safety-critical systems which must exhibit high transparency.
In this paper, to address this black-box limitation, we develop a simple analysis method which consists of 1) structural feature analysis: lists of the features contributing to inference process, 2) linguistic feature analysis: lists of the natural language labels describing the visual attributes for each feature contributing to inference process, and 3) consistency analysis: measuring consistency among input data, inference (label), and the result of our structural and linguistic feature analysis. 
Our analysis is simplified to reflect the actual inference process for high transparency, whereas it does not include any additional black-box mechanisms such as LSTM for highly human readable results.
We conduct experiments and discuss the results of our analysis qualitatively and quantitatively, and come to believe that our work improves the transparency of neural networks. 
Evaluated through 12,800 human tasks, 75\% workers answer that input data and result of our feature analysis are consistent, and 70\% workers answer that inference (label) and result of our feature analysis are consistent.
In addition to the evaluation of the proposed analysis, we find that our analysis also provide suggestions, or possible next actions such as expanding neural network complexity or collecting training data to improve a neural network.

\end{abstract}
  \vspace{-5mm}

\begin{figure}[ht]
  \vspace{5mm}
  \centering
    \subfloat[Input image]{
    \centering
    \includegraphics[scale=0.31]{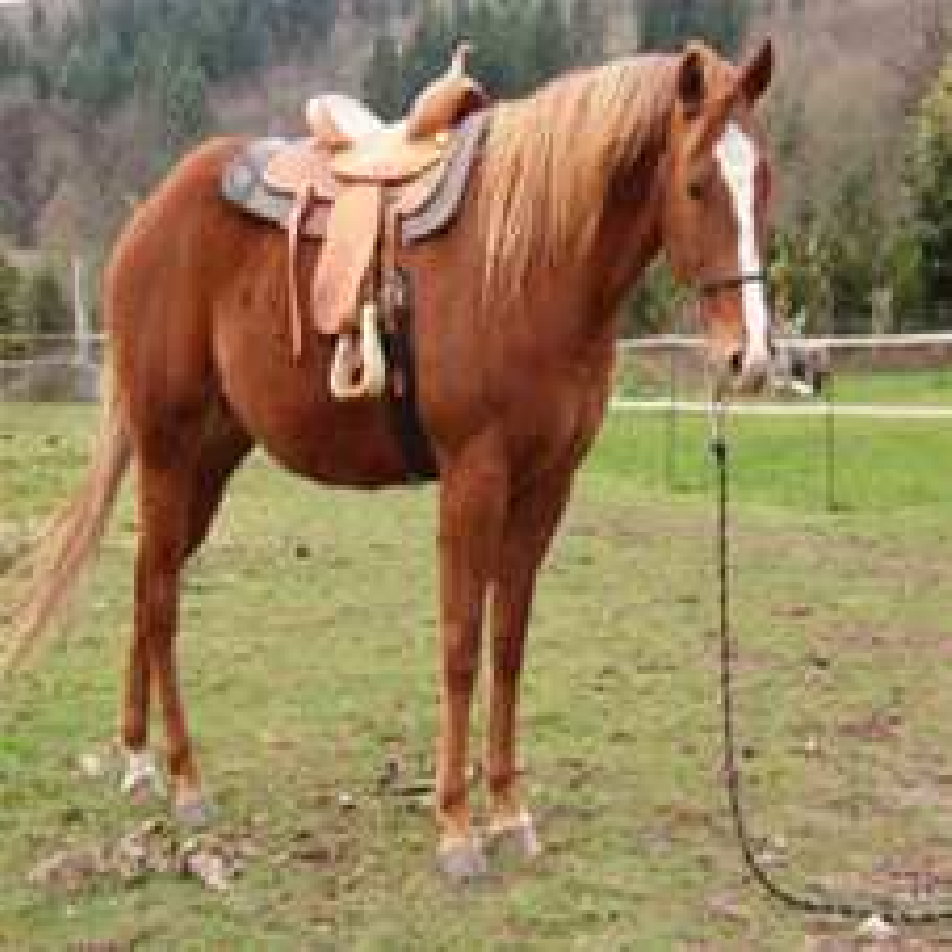} 
  \vspace{-1mm}
      }
      \hspace{5mm}
    \subfloat[Inference (label)]{
    \centering
    \includegraphics[scale=0.31]{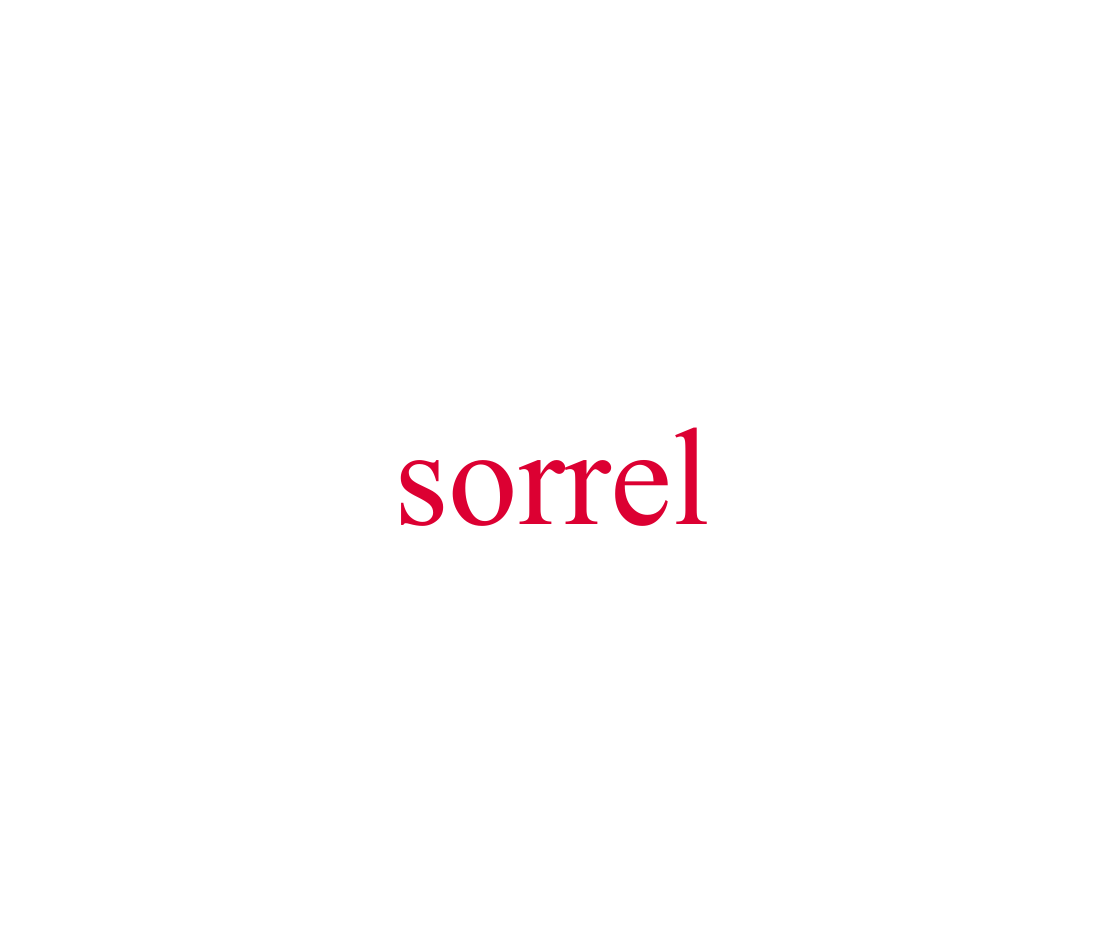} 
  \vspace{-1mm}
      } \\
    \subfloat[Structural \& linguistic feature analysis]{
    \centering
    \includegraphics[scale=0.31]{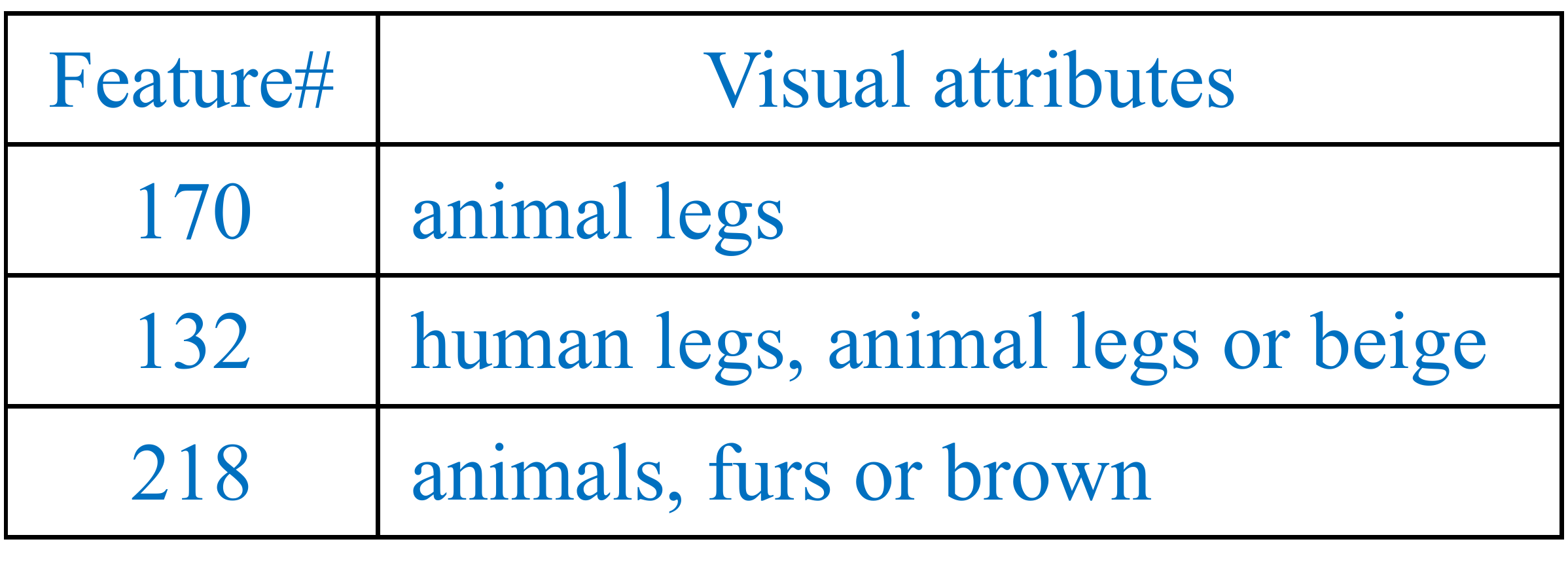} 
  \vspace{-1mm}
    \label{explanation_example}
  } \\
      \subfloat[Side information (optional)]{
    \centering
    \includegraphics[scale=0.31]{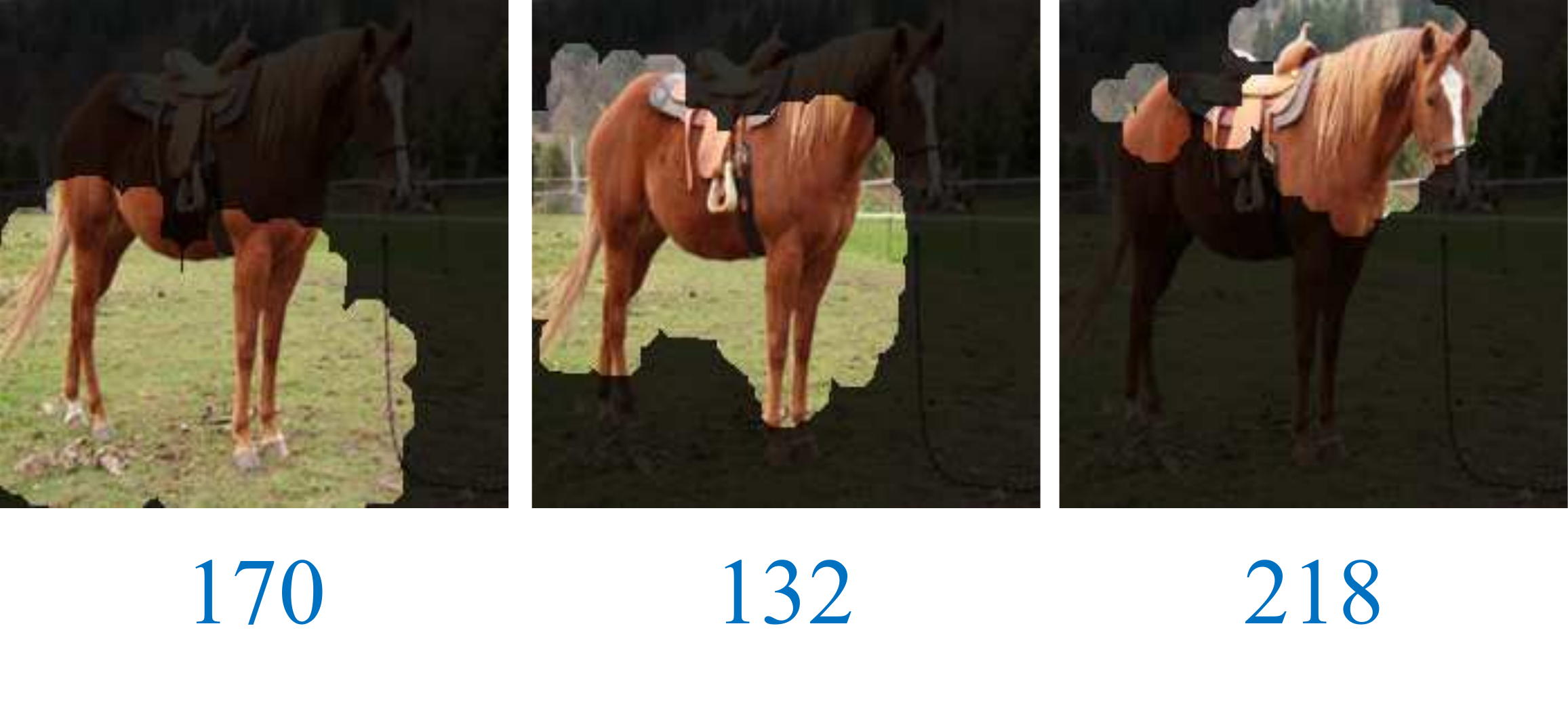} 
  \vspace{-1mm}
    \label{sideinfo_example}
  }
  \caption{Feature analysis example}
    \label{example}
\end{figure}

\section{Introduction}
\label{intro}
Machine learning techniques
such as deep neural networks has led to the widespread application
of systems that assign advanced environmental perception and decision-making to computer logics learned
from big data, instead of manually built rule-based logics~\cite{ILSVRC15,NIPS2013-5021,pennington2014glove,DBLP:journals/corr/abs-1708-02709,DBLP:conf/asru/GravesJM13,uchida}.
  Deep learning especially achieves unprecedented performance on several tasks. 
  For example, in the visual object recognition task outperformed humans~\cite{DBLP:conf/cvpr/2015}.

Machine-learning models are becoming indispensable components
even in systems that require safety-critical environmental
perception and decision making, such as automated driving systems~\cite{DBLP:journals/corr/BojarskiTDFFGJM16}.
To build high credibility for machine-learning models, 
  both high performance and transparency are important. 
In particular, safety-critical systems must exhibit transparency~\cite{dise1}. 
However, inference processes of machine-learning models
such as neural networks are considered as black boxes.
In this paper, a black box
refers to a situation, where, although feature activation can
be observed, the actual phenomenon cannot be understood.
In other words, machine-learning models show high performance but low transparency.
Thus, it is difficult for black-box deep learning to be applied to safety-critical systems 
such as automated driving in which the results of deep learning models 
can directly cause hazard~\cite{koopman2016challenges}.

Explainable AI (XAI) is a related research area which is focused and rapidly advanced recently~\cite{DBLP:journals/corr/Miller17a}.
There are studies in XAI that inference networks give human understandable explanations, as well as inference (label).
For example, image caption generation and visual explanation are problems to provide highly human understandable natural language descriptions.
  Caption generation is a verbalization method, which describes the objects and the circumstances happening in the input image
   by natural language sentences~\cite{DBLP:conf/icml/XuBKCCSZB15,DBLP:conf/cvpr/VinyalsTBE15}. 
  Visual explanations are generated by black-box explaining models such as LSTM,
   to explain rationales for classification decisions~\cite{DBLP:conf/eccv/HendricksARDSD16}.
  They generate highly human readable explanation, however by using mechanisms which do not reflect the actual inference process, because
  explanation generation and classification are done by different neural networks possibly sharing features, inference results (labels), etc.
  Even sharing features, explanation generation is done by black-box models (neural networks), and we cannot know they reflect the actual inference process.
  Inference networks which generate explanations have high performance but low transparency.

\begin{figure}
  \centering
    \includegraphics[width=84mm]{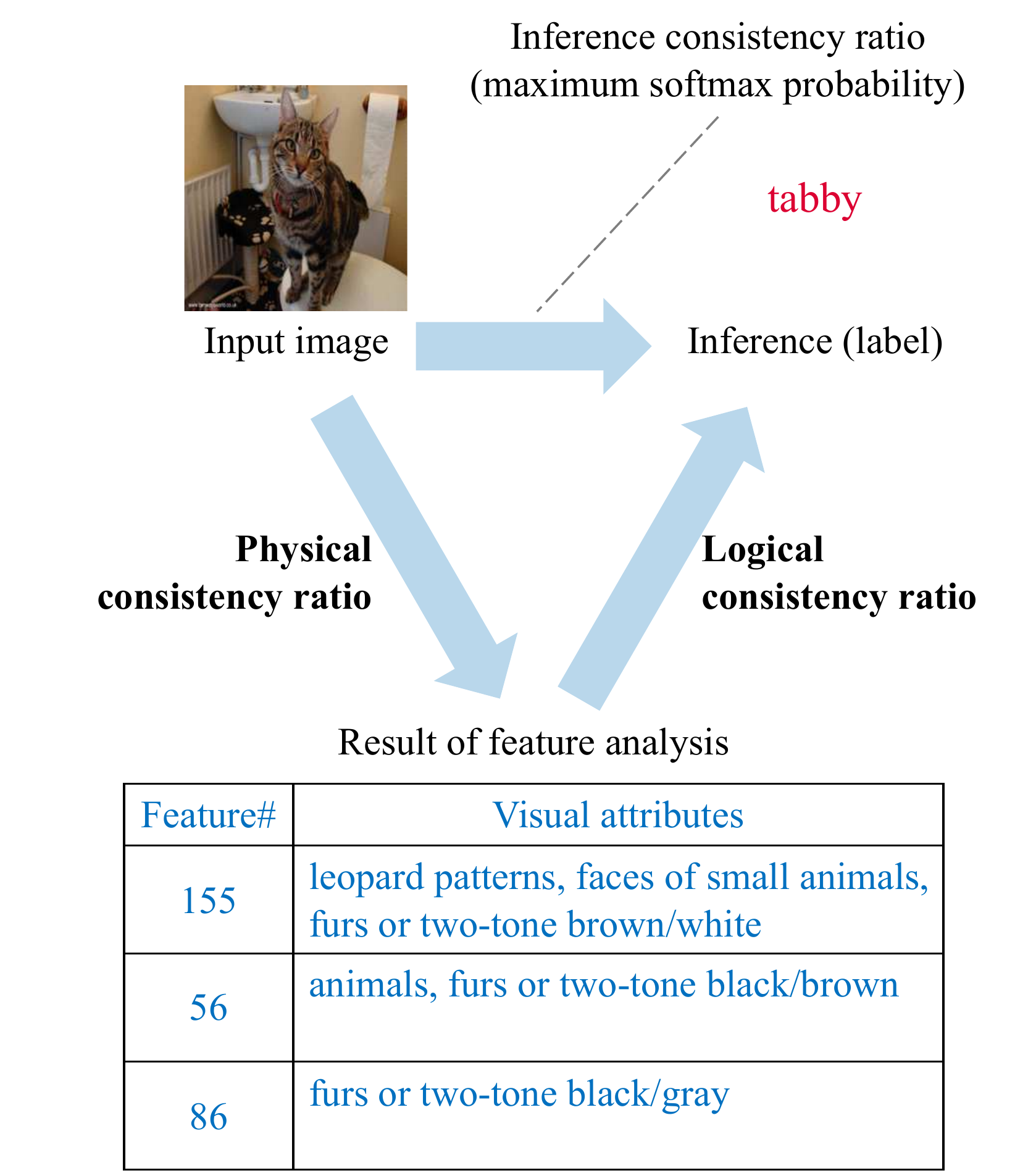} 
  \caption{Consistency analysis concept}
    \label{eval}
\end{figure}

In this paper, to address the black-box property of deep learning, 
we develop a simple analysis method which improves the transparency of inference processes of 
convolutional neural networks, hereinafter referred to as CNN~\cite{DBLP:journals/pr/FukushimaM82,NIPS2012-4824},
 as an example of deep learning models. 
We assume three types of analysis for inference process;
1) structural feature analysis, 2) linguistic feature analysis, and 3) consistency analysis. 
Results of structural feature analysis are lists of the features contributing to inference process.
The feature numbers are not human readable, but are useful when systems programmatically manage the inference process at test time. 
Results of linguistic feature analysis are lists of the natural language labels describing the visual attributes for each feature obtained through the structural feature analysis.
It is useful for humans to understand inference processes.
Figure \ref{example} is an example result of our feature analysis. The left and right columns in Fig. \ref{explanation_example} are 
structural feature analysis and linguistic feature analysis, respectively.
Consistency analysis is to measure the consistency among input data, inference (label), and result of feature analysis.
It is useful for discussion, such as identifying the cause of incorrect inference (label) and possible next actions to fix problems, etc.
Figure \ref{eval} shows the concept of consistency analysis. 
To show the usefulness of our proposed method, we conduct experiments including human evaluation, and have corresponding discussion on the experimental results.

This paper is an extended version of our previous workshop paper presented in Transparent and interpretable Machine Learning in Safety Critical Environments, NIPS2017 Workshop~\cite{timl}.

\section{Related Works}
DARPA started Explainable Artificial Intelligence program in 2017~\cite{XAI}.
It defines three approaches: 
Deep Explanation, Interpretable Models, and Model Induction.
The first and second are ex-ante approaches which design explainable features and explainable causal models {\bf in advance to training}.
The third is an ex-post approach which automatically derives new explainable models {\bf after training}. 
Our transparent analysis is an ex-post (after training) approach, but it does not derive new models and directly analyze the actual activation observed.

  Visualization of deep neural networks is an active study area recently~\cite{DBLP:journals/corr/GrunRNT16,DBLP:journals/corr/MontavonSM17}. Earlier studies are basically identify attention (focus) areas of input data 
  in receptive fields or heat maps~\cite{DBLP:journals/corr/ShrikumarGSK16,DBLP:conf/icann/BinderMLMS16,DBLP:journals/pr/MontavonLBSM17}. It indicates the areas in the input data 
  which the model is looking at~\cite{DBLP:journals/corr/BojarskiCCFJMZ16} during test time. 
Attention area of an input image is the very beginning part of CNN inference process, and is revealed by visualization methods. 
On the other hand, in this paper, we would like to provide analysis not only for input data, but also for inference process of neural networks.
We exploit receptive fields as side information indicating the locations of the visual attributes in input data.

  There is another type of works focusing on the visual attributes and intermediate features, {\it i.e.}, activation of neural network nodes. 
  One of past works analyzed the visual attributes for each node, 
  and it was revealed that low level attributes, such as black, brown, and furry are associated to neural network nodes~\cite{DBLP:conf/cvpr/EscorciaNG15}. 
  Another work interprets receptive fields as with visual attributes of neural networks, 
  and quantified the interpretability by using the number of human interpretable visual semantic concepts 
  learned at each hidden layer~\cite{bau2017netdissect}. 
  Among visualization techniques, in this paper, we use visual attributes for our transparent analysis.

  Pointing and Justification-based Explanation (PJ-X) is one of the latest explanation methods in XAI~\cite{DBLP:journals/corr/ParkHASDR16}. 
It can provide highly human understandable explanation, by attention areas of input data space as introspective explanations (true explanation) and justification explanations at the same time. The former provides explanations of the input space, but does not provide analysis for the inference processes. The latter does not address the black-box property of target models, because it uses another black-box method LSTM to generate the explanation. PJ-X does not analyze the relationship between the inference results and features of neural networks, and introducing an additional black-box model for explanation cannot address transparency. Therefore, the purpose of PJ-X is not an analysis for improving transparency of deep neural inference process.

\section{Observation of Feature Contribution}
We observed {\tt conv5} feature of CaffeNet~\cite{ding2014theano,NIPS2012-4824} on selected ImageNet training data, to understand the behavior of features. 
Although ImageNet has approximately $1300$ training images per class, for simplicity, we selected $100$ examples for each class, with top-$100$ softmax probability on the ground truth classes. 

We first make a natural assumption that inference (label) is based not on inactive features, but on highly activated features, and derived the following assumption.
\begin{itemize}
\item {\bf Assumption 1}. Features highly activated in the inference process have contributions to inference (label).
\end{itemize}
 This assumption applies especially for ReLU, which CaffeNet uses as activation functions, because ReLU is a half-linear positive monotonic function.

\subsection{Magnitude of Feature Activation}
\begin{figure}[ht]
  \centering
  \includegraphics[width=84mm]{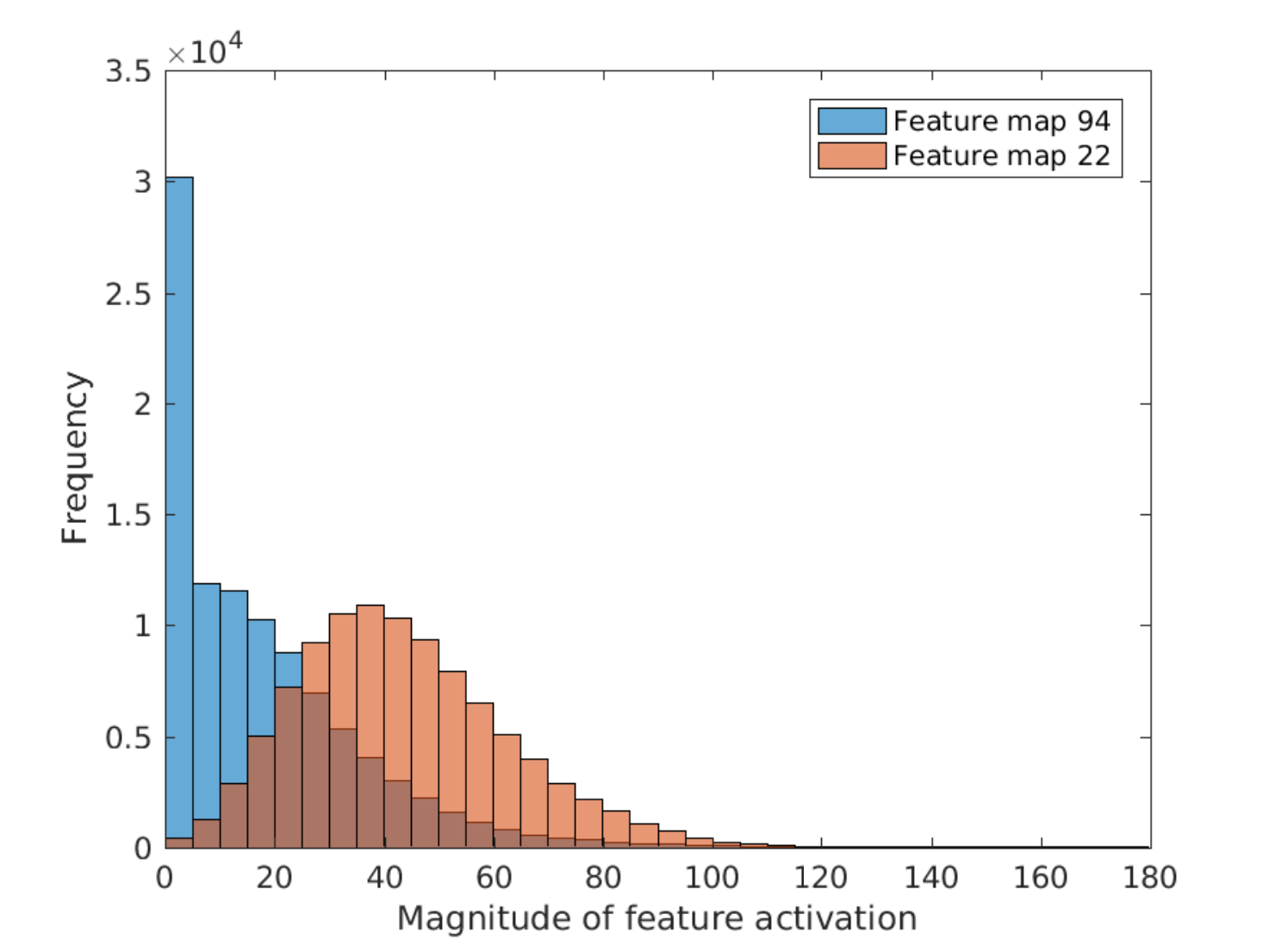}
  \caption{Dynamic ranges of different feature maps' activation. The mode activation on feature map $94$ is $0$, whereas that on feature map $22$ is $38$. The shapes of distribution are also different.}
  \label{dynamic_range}
\end{figure}

Then we look into activation on each feature map in {\tt conv5}, and found that the magnitude of activation changes for different features. Therefore, definition of high activation varies depending on features. Figure \ref{dynamic_range} shows the histograms of activation on example feature maps $94$ and $22$, which have the smallest and largest mean values, respectively. The modes of activation magnitude are different each other.
These distributions are not gaussian, because negative values are cancelled by ReLU activation function.
It is clear that the distributions of activation on feature maps $94$ and $22$ are different. 
By this analysis, we derived the following assumption.
\begin{itemize}
\item {\bf Assumption 2}. Activation in different features have different dynamic ranges.
\end{itemize}

\subsection{Features and Visual Attributes}
\begin{figure}[t]
  \centering
  \vspace{10mm}
    \subfloat[Feature map 226]{
    \centering
    \footnotesize
    \stackunder[5pt]{\includegraphics[scale=0.4]{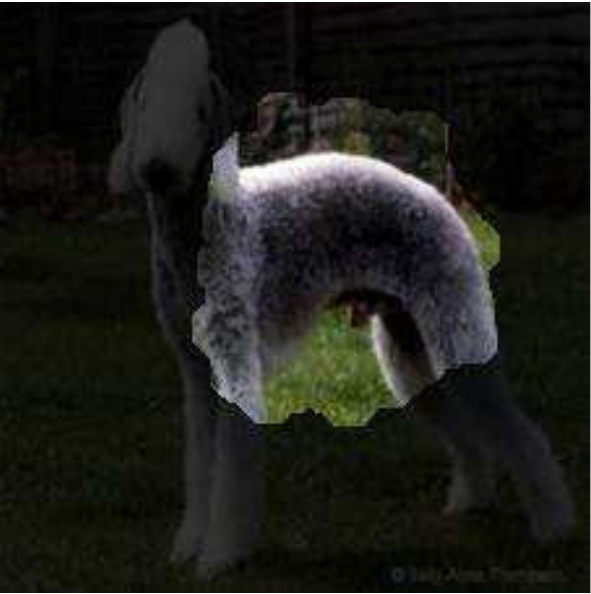}}{furly}%
     \hspace{12mm}
    \stackunder[5pt]{\includegraphics[scale=0.4]{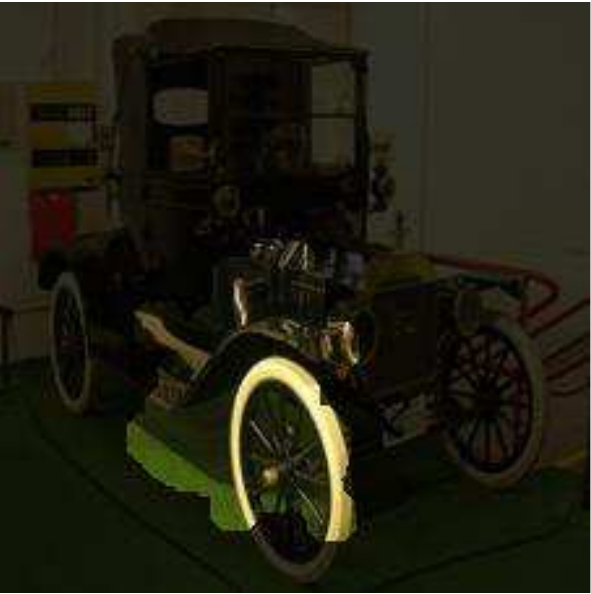}}{rubber tires}%
  } \\
    \subfloat[Feature map 230]{
    \centering
    \footnotesize
    \stackunder[5pt]{\includegraphics[scale=0.4]{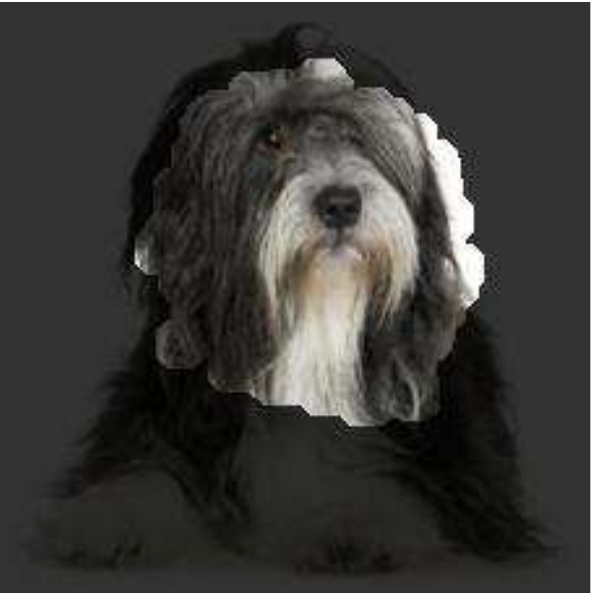}}{furly}%
     \hspace{12mm}
    \stackunder[5pt]{\includegraphics[scale=0.4]{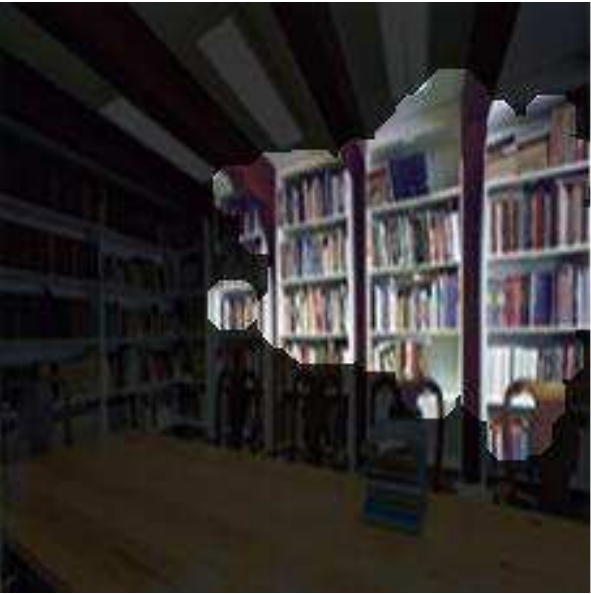}}{fine cell patterns}%
  }
  \caption{Visual attributes associated with features. Left column: furly; right column: rubber tires and fine cell patterns visual attributes appear on feature maps $226$ and $230$.}
    \label{manytomany}
\end{figure}

Figure \ref{manytomany} includes three visual attributes: furly, rubber tires, and fine cell patterns, but only two feature maps $226$ and $230$. These features share the visual attribute furly, while at the same time, have the other visual attributes different each other. 
This observation implies the following assumption.
\begin{itemize}
\item {\bf Assumption 3}. Visual attributes and features are in a many-to-many relationship.
\end{itemize}

\section{Proposed Analysis}
In this section, we propose a transparent analysis method to improve the transparency of deep neural inference processes based on assumptions.
We carry out both training time and testing time feature analysis to obtain three types of features, as described in \ref{sec_feat_anal}. 
We performed manual feature annotation to associate features with visual attributes, as described in \ref{feat_annotation}.
Then, three types of consistency ratios among input image, result of our proposed feature analysis, and inference (label) are measured through human tasks, as described in \ref{consist}.

\subsection{Structural Feature Analysis}
\label{sec_feat_anal}
\begin{figure*}[ht]
  \centering
  \includegraphics[scale=0.34]{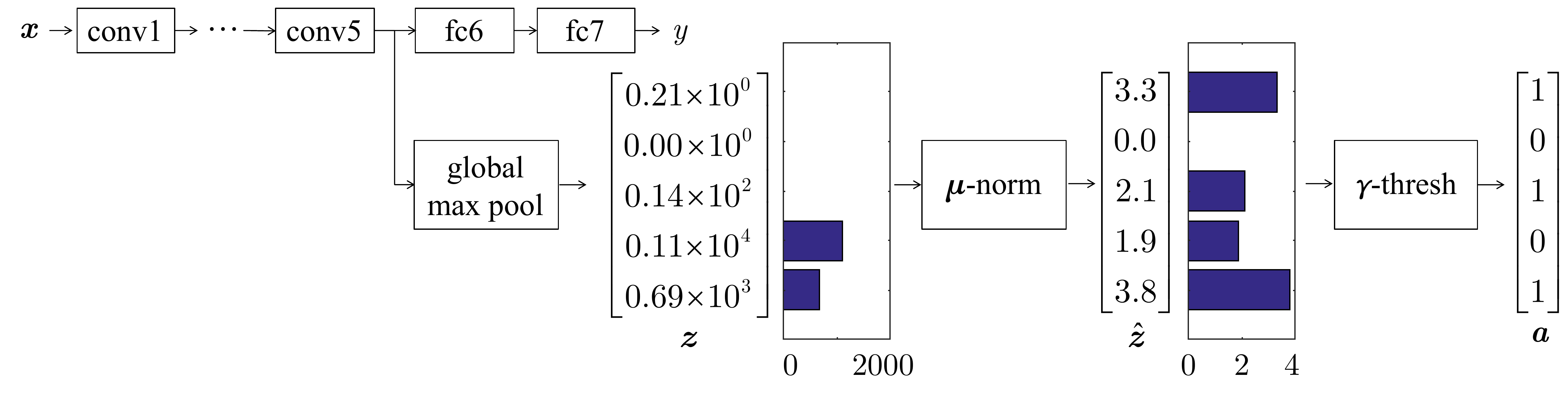}
  \caption{Activated feature}
  \label{act_feat}
\end{figure*}

\begin{figure}[ht]
  \centering
  \includegraphics[scale=0.34]{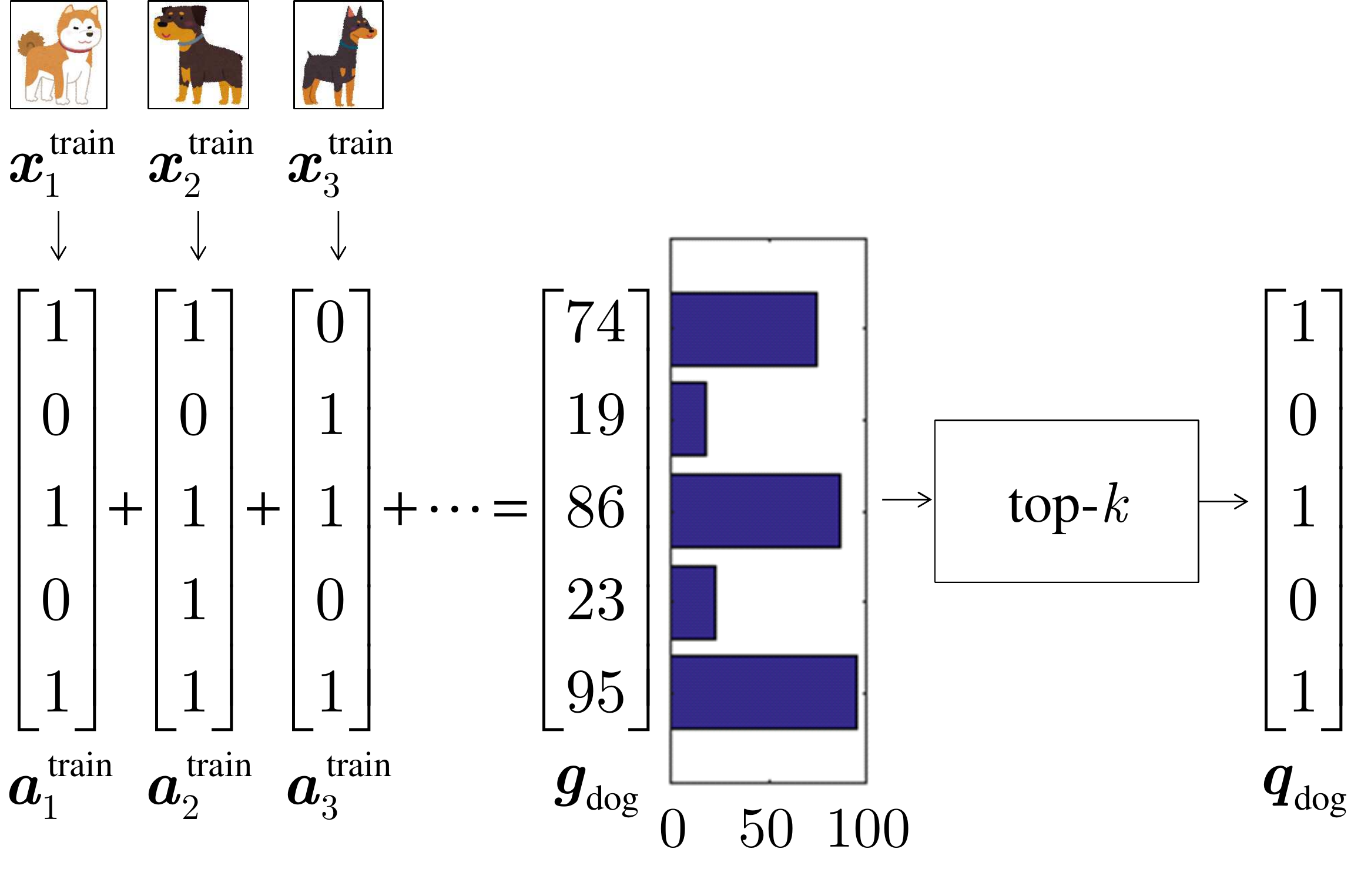}
  \caption{Class frequent feature}
  \label{freq_feat}
\end{figure}

\begin{figure}[ht]
  \centering
  \includegraphics[scale=0.34]{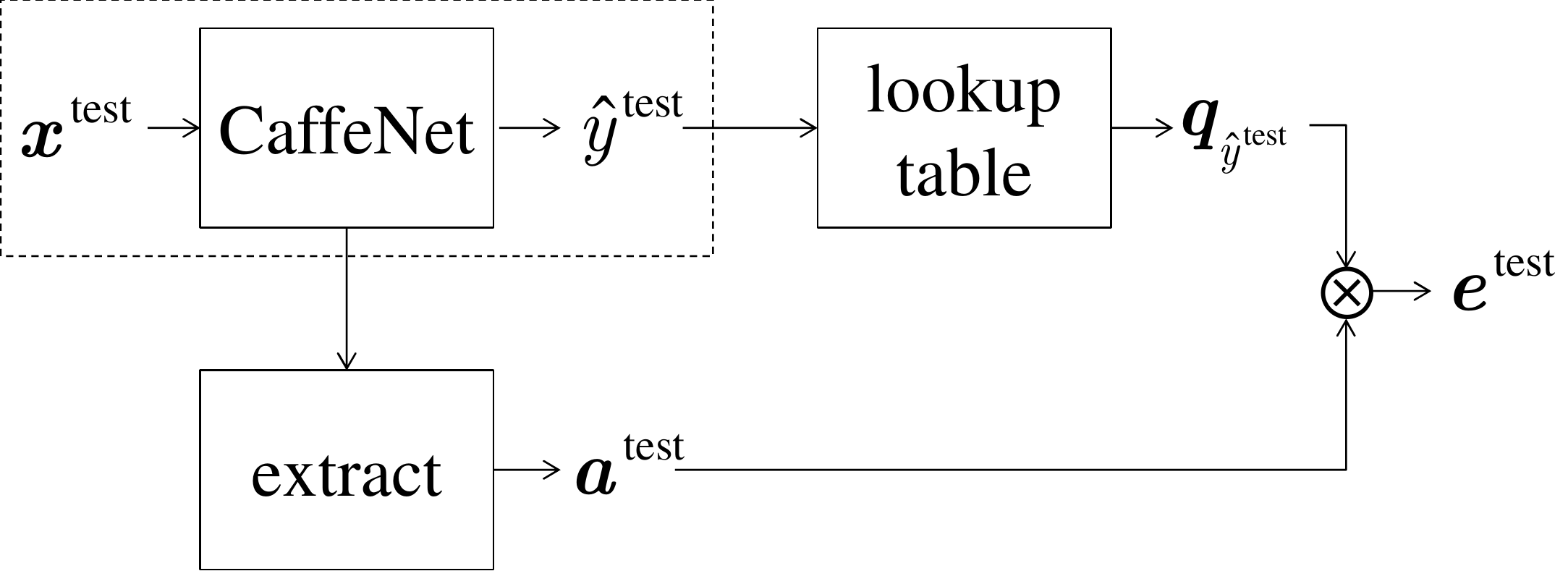}
  \caption{Inference feature}
  \label{expl_feat}
\end{figure}

We propose three concepts of features; 1) activated feature, 2) class frequent feature, and 3) inference feature as depicted in Fig. \ref{act_feat}, \ref{freq_feat}, and \ref{expl_feat}. 
Activated feature and inference feature are defined for each inference, whereas class frequent feature is defined for each class.


To analyze inference process, we focus on the activation of an intermediate feature called {\tt conv5} which is the final convolved feature in CaffeNet. 
It is reported that {\tt conv5} of AlexNet, which is also the final convolution layer, learns high level visual concepts such as objects and parts, and they are interpretable for humans~\cite{bau2017netdissect}.
Let $\bm{x}$ and $y$ are the input and the output of CaffeNet. Specifically $\bm{x}_i^{\text{train}}$, $y_i^{\text{train}}$ and $\bm{x}^{\text{test}}$ are those of the training data and the testing data.

\textbf{Activated feature} $\bm{a}$ in Fig. \ref{act_feat} is the binarized feature vector generated from {\tt conv5}. 
Activated feature $\bm{a}$ is a binary feature vector, however, CNN feature maps have spacial dimension. We decided to ignore the location of activation for simplicity, and applied global max pooling to contract {\tt conv5}, which is originally $13 \times 13 \times 256$ tensor, into 256-dimensional feature vector $\bm{z}$, as it is the simplest way to obtain a vector from a tensor. 
Therefore, we consider a feature map, with spacial feature elements, as a single feature.
The element of the vector ${\bm a}$ is one if the associated feature, {\it i.e.}, the feature map in {\tt conv5}, is activated.
Based on {\bf Assumption 2}, in order to judge weather the feature is activated or not, it is necessary to use statistical information such as mean, variance, or higher moment to capture the differences among features. In this paper, we decided to use mean normalization and thresholding. 
We compute a mean-normalized feature vector $\hat{\bm{z}}$ from a feature vector $\bm{z}$, as each element of $\bm{z}$ has varying dynamic range, and normalization makes them comparable each other. Thresholding $\hat{\bm{z}}$ at $\gamma$ gives a binarized feature vector $\bm{a}$ corresponding to $\bm{x}$.

\textbf{Class frequent feature} $\bm{q}$ in Fig. \ref{freq_feat} is binary vectors indicating the frequently activated features for each class. 
We hypothesize that each class has a different frequent activation pattern which is obtained by the following procedure.
Fig. \ref{freq_feat} shows how to compute the class frequent feature for an example class: dog. The training data $\bm{x}_{i}^{(\text{train})}$ of the dog class is binarized into $\bm{a}_{i}^{(\text{train})}$, and their summation over $i$ counts how many times each feature is activated for the dog class in the training data. After summation, we select the top-$k$ frequent features which consist the class frequent features for the dog class, where $k = 3$ in the case of Fig. \ref{freq_feat}. Class frequent features are computed for each class at the training time, and stored in a lookup table to be used in the testing time, like $\bm{q}(\text{dog}) = [1,0,1,0,1]$, $\bm{q}(\text{cat}) = [1,1,0,0,1]$ and $\bm{q}(\text{bird}) = [1,0,0,1,1]$. 

\begin{figure}
  \centering
  \includegraphics[width=84mm]{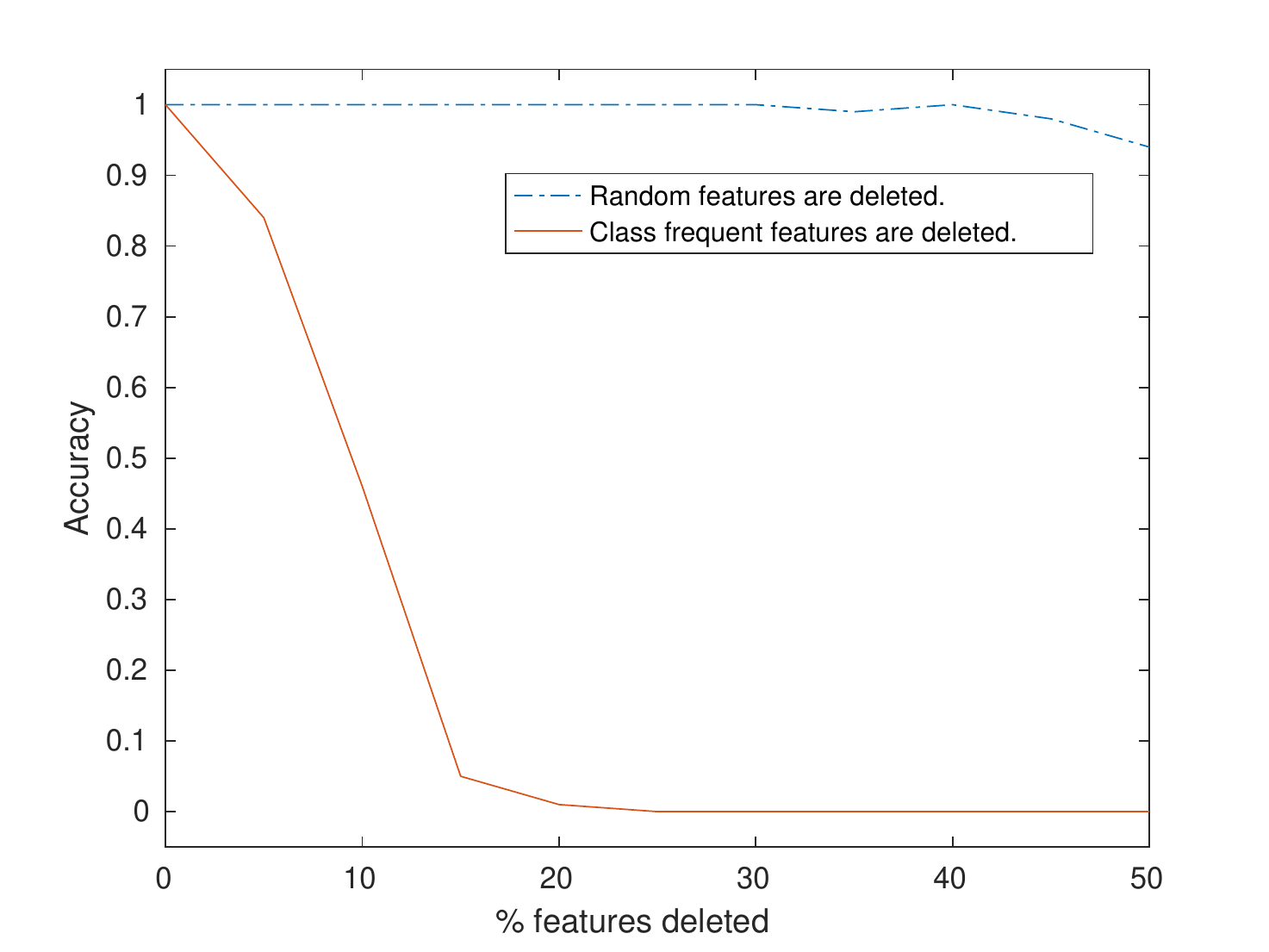}
  \caption{Accuracy decay with feature map deletion}
  \label{feat-delete}
\end{figure}

To check the validity of class frequent features,
we made two CaffeNets whose randomly selected feature maps in {\tt conv5}, and frequently activated feature maps in {\tt conv5}, are replaced by zero, respectively. 
Figure \ref{feat-delete} shows how CNN accuracy for a sample class decays when we delete random feature maps or the class frequent feature maps for the class. 
We see fast accuracy decay when the deleted feature maps are frequently activated for the class. 
On the other hand, the convolutional neural network is robust against deleting randomly selected feature maps.
Original CaffeNet probably has redundant feature maps in {\tt conv5}.
This observation shows that class frequent features play important rolls in inference. 

\begin{figure*}[ht]
  \centering
  \includegraphics[scale=0.45]{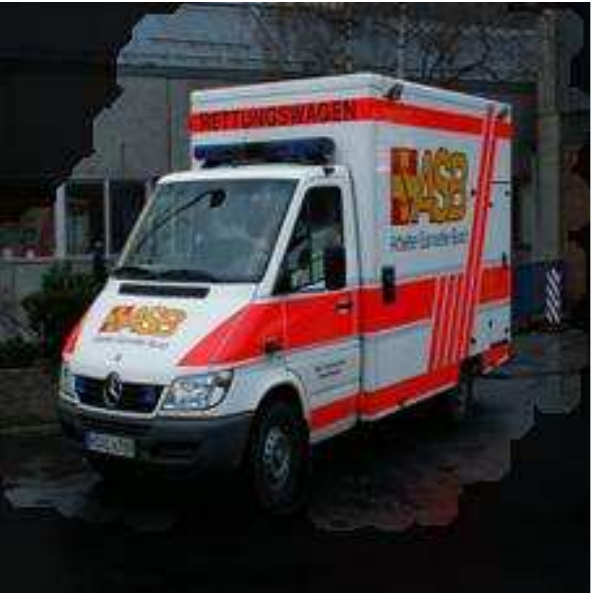} \hspace{5mm}
  \includegraphics[scale=0.45]{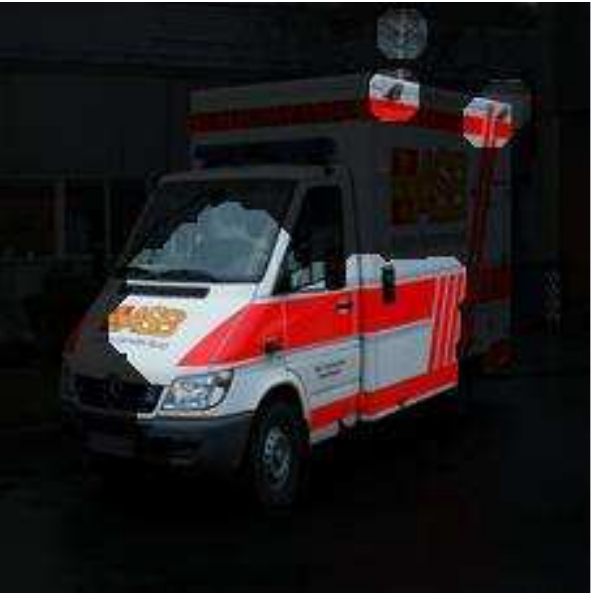} \hspace{5mm}
  \includegraphics[scale=0.45]{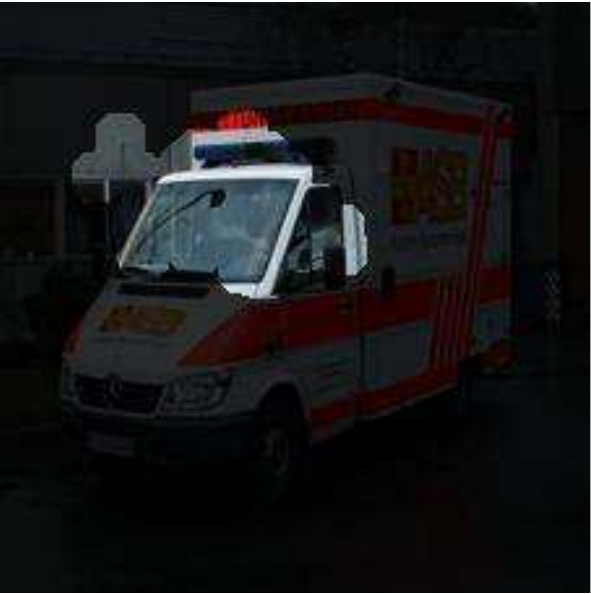} \hspace{5mm}
  \includegraphics[scale=0.45]{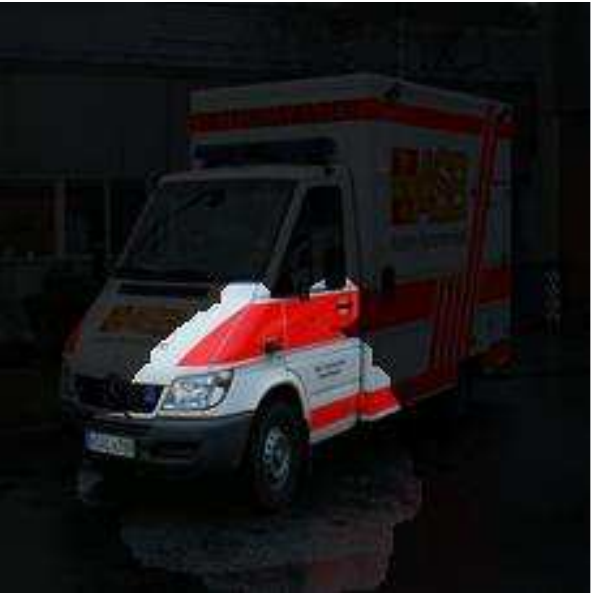} \hspace{5mm}
  \includegraphics[scale=0.45]{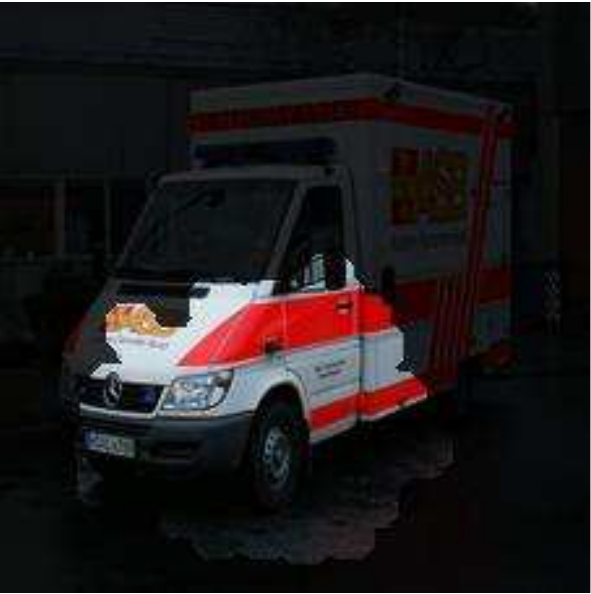}
  \caption{Receptive fields of feature maps included in the class frequent feature for ambulance class. Left to right: receptive fields of feature maps $084$, $177$, $234$, $239$, and $242$.} 
  \label{ambulance}
\end{figure*}

To understand the relationship between class frequent features and inference (label), 
we display the areas on which the active elements on the feature maps included in a class frequent feature focus. 
The ambulance class has the class frequent feature including feature maps $084$, $177$, $234$, $239$, and $242$, and the corresponding receptive fields are visualized in Fig. \ref{ambulance}. 
Receptive fields are generated for each feature map, which we call the target feature map below.
For simplicity, we generated receptive fields by 1) replacing the elements on the target feature map with the activated features, {\it i.e.}, binarizing the target feature map, 2) replacing the off target feature maps (any feature maps except feature map $084$ if we generate receptive fields for the feature map $084$) with zero, 3) back propagating the modified feature including both target and off target feature maps to the input space with unpooling using the stored pooled location on max pooling layers, and 4) post processing including image binarization and dilation by a disk shape.
Feature maps $084$ and $177$ respond the white-red (or orange) two tone color; feature map $234$ responds windows; and feature maps $239$ and $242$ respond tires in Fig. \ref{ambulance}.
We observe key parts of ambulance vehicles in the receptive fields. It is suggested that deep neural inference process is based on these key parts, and we derived the following assumption.
\begin{itemize}
\item {\bf Assumption 4}. Features frequently activated for the class of inference (label) have high contributions to inference (label).
\end{itemize}

\textbf{Inference feature} $\bm{e}^{\text{test}}$ in Fig. \ref{expl_feat} is the overlap between the activated feature $\bm{a}^{\text{test}}$ and the class frequent feature $\bm{q}^{\text{test}}$ for a single test data point $\bm{x}^{\text{test}}$, where $\otimes$ denotes element-wise product. Based on {\bf Assumption 1} and {\bf Assumption 4}, features contributing to inference process should be a part of both activated feature and class frequent feature. The dotted box in Fig. \ref{expl_feat} is the conventional inference without feature analysis. $\bm{a}^{\text{test}}$ is computed based on $\bm{x}^{\text{test}}$, whereas the class frequent feature $\bm{q}^{\text{test}}$ is just lookup by the inference (label) $y^{\text{test}}$ given by CaffeNet, because the ground truth is unknown in the testing time. $\bm{e}^{\text{test}}$ is the result of structural feature analysis.
The number of feature vector elements in an inference feature $\bm{e}^{\text{test}}$ is generally variable for each inference. Due to the human readability, in an inference feature $\bm{e}^\textrm{test}$, we show at most top-$\ell$ feature vector elements with the maximum mean-normalized activation. 

\subsection{Linguistic Feature Analysis}
\label{feat_annotation}
To generate human readable analysis, we annotate visual attributes for each feature by looking at the input samples on which it is activated in the focused network. Although there are many ways to achieve human readable visual attributes, we decided to conduct human annotation, because it is the most simple method. 

\begin{figure}[t]
  \centering
    \includegraphics[width=84mm]{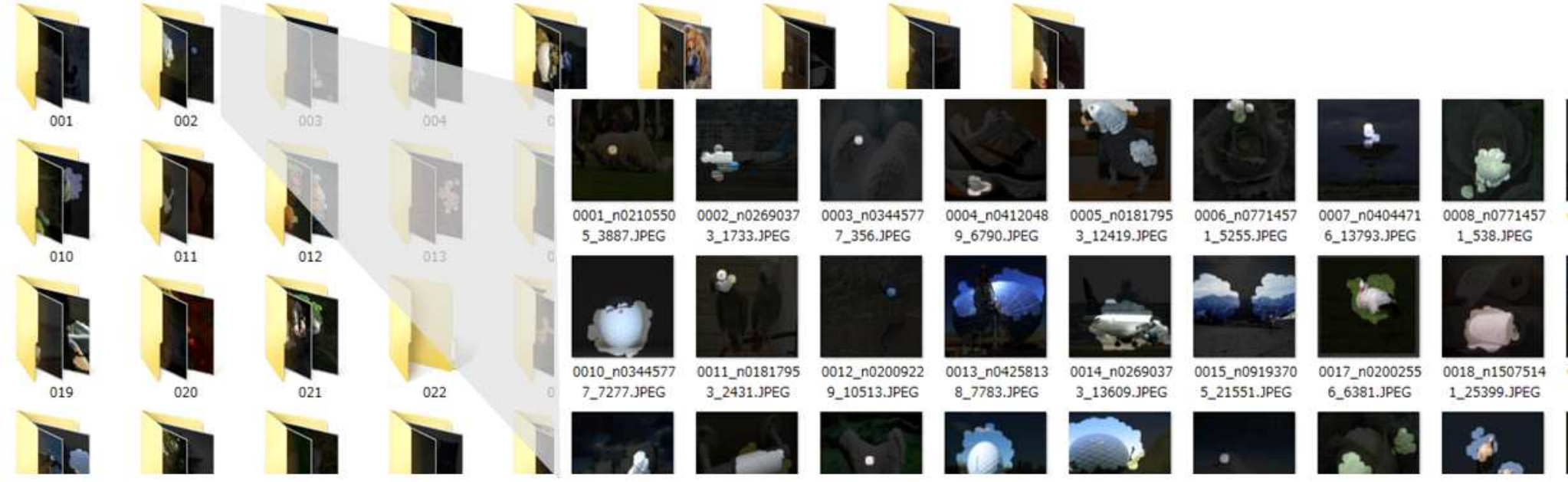} 
  \caption{Sample images and receptive fields for feature annotation. Each feature has a folder, and a folder has images with high activation on it. Workers see receptive fields in a folder, and annotate them.}
    \label{anno_ex}
\end{figure}

{\bf Annotation Data} is prepared by using the training data set. At first, we select a subset of training data suitable to feature annotation. And then, for each feature, we sample images so that their inference features (identified by the ground truth labels) include it. We generated receptive fields for the human annotator to understand the part of the image where visual attributes appear, as shown in Fig. \ref{anno_ex}. 

\begin{figure}[t]
  \centering
    \includegraphics[width=84mm]{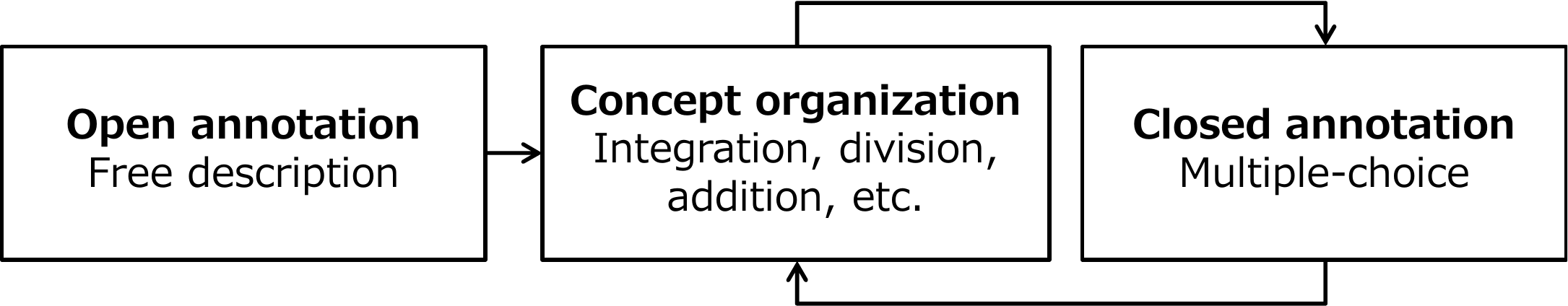} 
  \caption{Feature annotation process}
    \label{anno_process}
\end{figure}

{\bf Annotation Process} is to iterative way to annotate the combination of multiple features representing a single visual attribute, and vice versa.
In order to annotate this many-to-many relationship based on {\bf Assumption 3}, starting from free description, feature annotation is repetitively refined. We defined a process which consists of three steps; 1) open annotation, 2) label organization, 3) closed annotation.
The open annotation is the first step where a human annotator annotates all features by free description. Human annotation may have some fluctuation at this step. Second in the label organization step, similar visual attributes are integrated, different visual attributes with the same label are divided, new visual attributes are introduced, etc., so that the fluctuated labels of feature annotation are well organized. Then, the human annotator works on the closed annotation to classify features into a set of visual attributes, {\it i.e.}, multiple answers allowed, from the all visual attributes defined in the last step. Label organization and closed annotation steps are repeated to refine the feature annotation, as depicted in Fig. \ref{anno_process}.

\subsection{Consistency Analysis}
\label{consist}
To gain further insight, we measure the consistency among input data, inference (label), and result of our proposed feature analysis, {\it i.e.}, inference feature. 
This measurement is for discussion, checking whether our analysis method or the target neural network are incorrect when we get incorrect analysis, identifying possible next actions to fix problems, etc.

We propose \textbf{physical consistency ratio} (PCR) and \textbf{logical consistency ratio} (LCR), 
which are the consistency between inference feature and input data, and consistency between inference feature and inference (label), respectively (Fig. \ref{eval}). 
These two ratios are measured through human tasks.
In addition to two measures for consistency, we use the softmax probability corresponding to the class of inference (label), {\it i.e.} the maximum softmax probability, 
as \textbf{inference consistency ratio} (ICR), the consistency between input data and inference (label). All the ratios are in the range of $0.0$ to $1.0$.

\section{Experiments}
In this section, we conducted experiments to test our proposed analysis method.
We try to analyze the inference processes of the publicly available CaffeNet with the weighs pre-trained on ImageNet.
These feature vectors were binarized by the method introduced above, with the binarization threshold $\gamma = 2$. We chosen $k = 5$ the number of feature vector elements in a class frequent feature, and $\ell = 3$ the maximum number of feature vector elements in an inference feature.
Receptive fields for each feature vector elements in inference features are accompanied with the result of feature analysis as informative clue for human feature annotation, and side information to support the analysis. 

As with the feature analysis, selected $100$ training images per class are used for computing mean values for each feature map in {\tt conv5}, class frequent features, and annotating visual attributes by human. 
On the other hand, we reduced the $1,000$ object categories of ImageNet to $32$ for testing, because it is difficult for human to distinguish $1,000$ categories and understand the corresponding analysis precisely. The $32$ classes are a subset of ImageNet $1,000$ classes, which are programmatically selected according to the WordNet~\cite{Miller1990} hierarchy, such that each new class has approximately the same number of WordNet synsets. 

Human evaluation was done on Amazon Mechanical Turk. 
For each input image, we made two questions for the physical consistency and logical consistency on our feature analysis.
\begin{enumerate}
\item Is the inference feature relevant to the whole or parts of the input image? 
\item If an object satisfies the inference feature, is it an object in the class of inference (label)? 
\end{enumerate}
The first question was asked {\it without} showing inference (label), and the second one was asked {\it without} showing the input image.
The list of response alternatives shown to workers were strongly agree, agree, disagree, and strongly disagree. 
After obtaining the results from workers, we merged the former two and the latter two into agree and disagree, respectively.
The results of the questions are used to evaluate physical consistency ratio and logical consistency ratio, respectively.
Each question is redundantly assigned to discrete workers to eliminate individual biases, and the averaged ratios are in the range of $0.0$ (all workers disagree) to $1.0$ (all workers agree). 
To evaluate these ratios, we conducted $12,800$ human tasks for total (Table \ref{HIT}).
We also recorded the softmax probability of the class of inference (label), as inference consistency ratio. 
\begin{table}[ht]
  \caption{Number of human tasks to evaluate consistency measures. Redundancy added by discrete samples and workers is to eliminate individual biases. Inference consistency ratio (maximum softmax probability) is automatically computed.}
  \label{HIT}
  \centering
  \begin{tabular}{ |c||c|c|c||c|  }
 \hline
 Measure & Class & Sample & Worker & Total tasks \\ [0.5ex] 
 \hline\hline
PCR & 32 & 10 & 20 & 6,400 \\
LCR & 32 & 10 & 20 & 6,400 \\
 \hline
\end{tabular}
\end{table}

\begin{figure*}[ht]
  \centering
  \vspace{10mm}
  \subfloat[Distribution on correct inference]{
    \captionsetup{font={scriptsize}}
    \centering
    \includegraphics[scale=0.6]{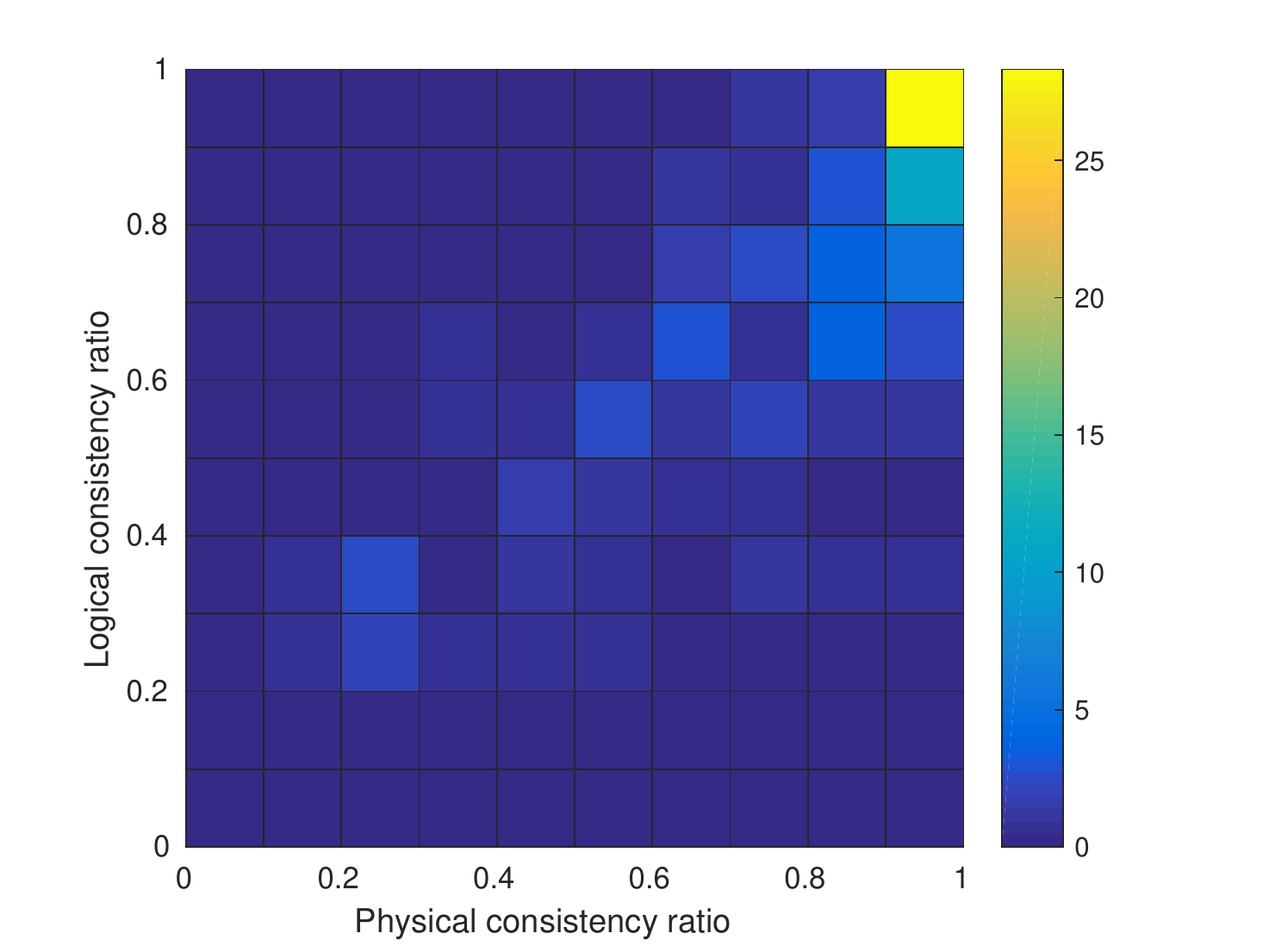} 
  \label{correct_pdf2d}
  } \hspace{-11mm}
    \subfloat[Distribution on incorrect inference]{
    \captionsetup{font={scriptsize}}
    \centering
    \includegraphics[scale=0.6]{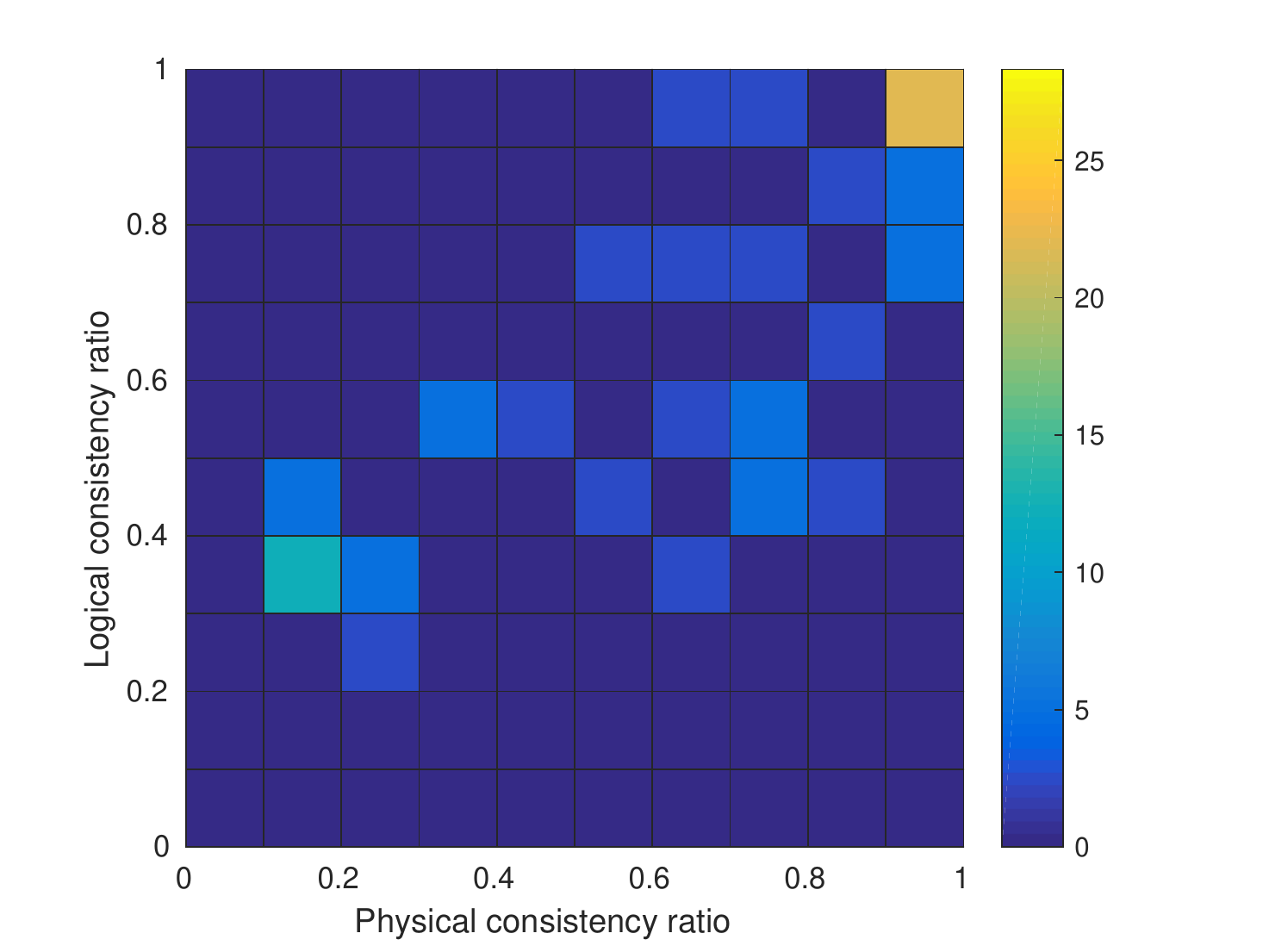} 
  \label{incorrect_pdf2d}
  }
  \caption{Joint discrete probability distribution for physical consistency ratio and logical consistency ratio.}
  \label{ratios_result}
\end{figure*}

Figure \ref{correct_pdf2d} and \ref{incorrect_pdf2d} show the joint discrete probability distribution between 
physical consistency ratio and logical consistency ratio on correct inference and incorrect inference, respectively.
When inference is incorrect, the peak is low although it is located in $(0.8,1.0)$, and the both physical consistency ratio and logical consistency ratio spread throughout from low consistency to high consistency.
On the other hand, both ratios tend to be high when inference is correct. 
The distribution on correct inference is clearly high contrast compared to that on incorrect inference.
Therefore, our method provides better analysis for correct inference than that for incorrect inference. 
The mean values of physical consistency ratio, logical consistency ratio, and inference consistency ratio, {\it i.e.}, softmax probability, over entire experimental data even including incorrect inference were $0.75$, $0.70$, and $0.48$, respectively.
According to these results, our method gained consensus on humans, overall distribution of consistency ratios are reasonable.

\section{Discussion}
In this section, we show whether our proposed simple analysis improves the transparency of inference processes of convolutional neural networks, 
and we study what practical discussion in a machine learning training and testing process can be done on a neural network thanks to that improved accuracy.

\begin{figure*}
  \centering
  \subfloat[Analysis results with low ICR]{
    \captionsetup{font={scriptsize}}
    \centering
    \includegraphics[scale=0.1,bb=0 0 2220 2703]{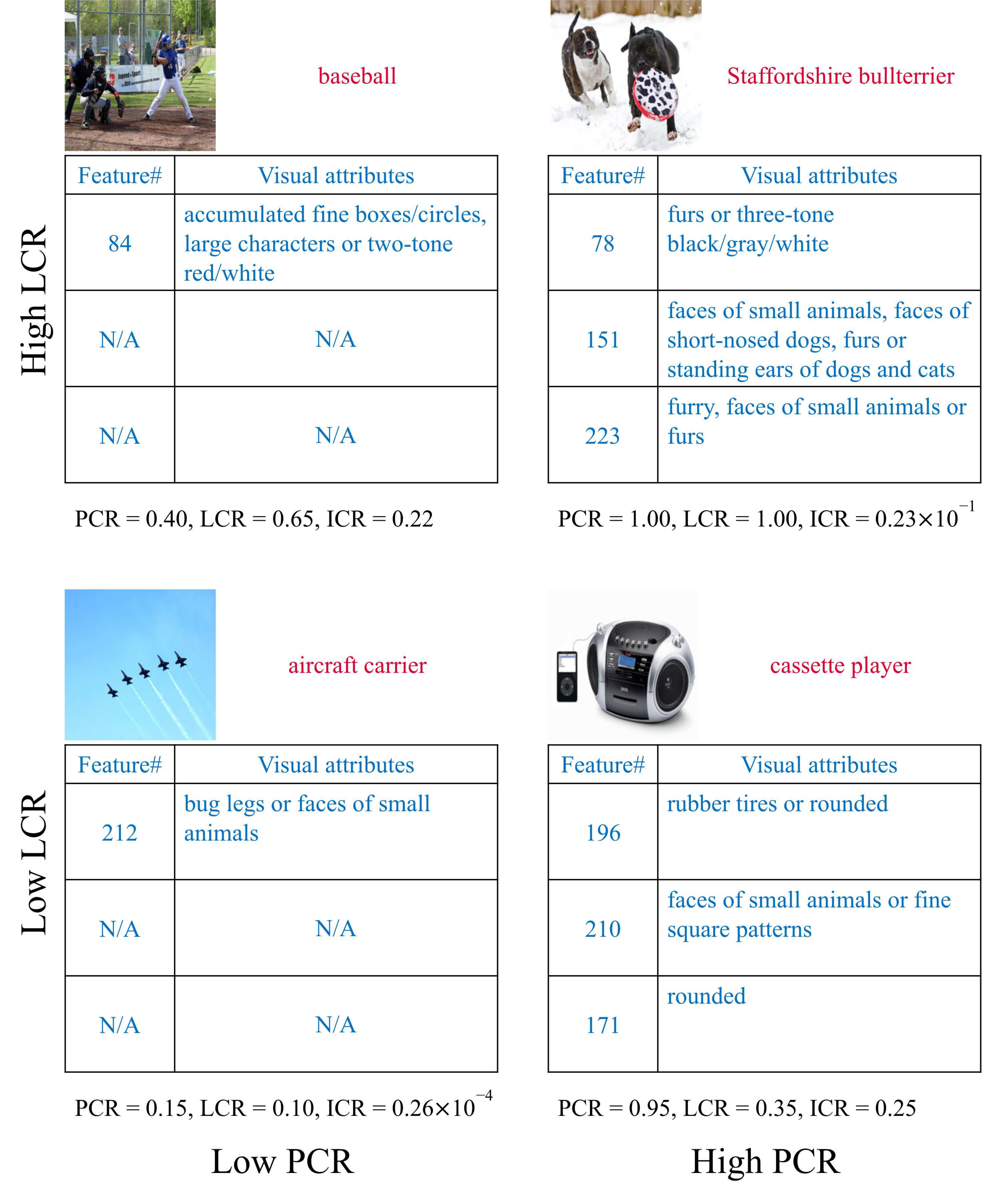} 
  \label{low_icr}
  }
  \subfloat[Analysis results with high ICR]{
    \captionsetup{font={scriptsize}}
    \centering
    \includegraphics[scale=0.1,bb=0 0 2219 2703]{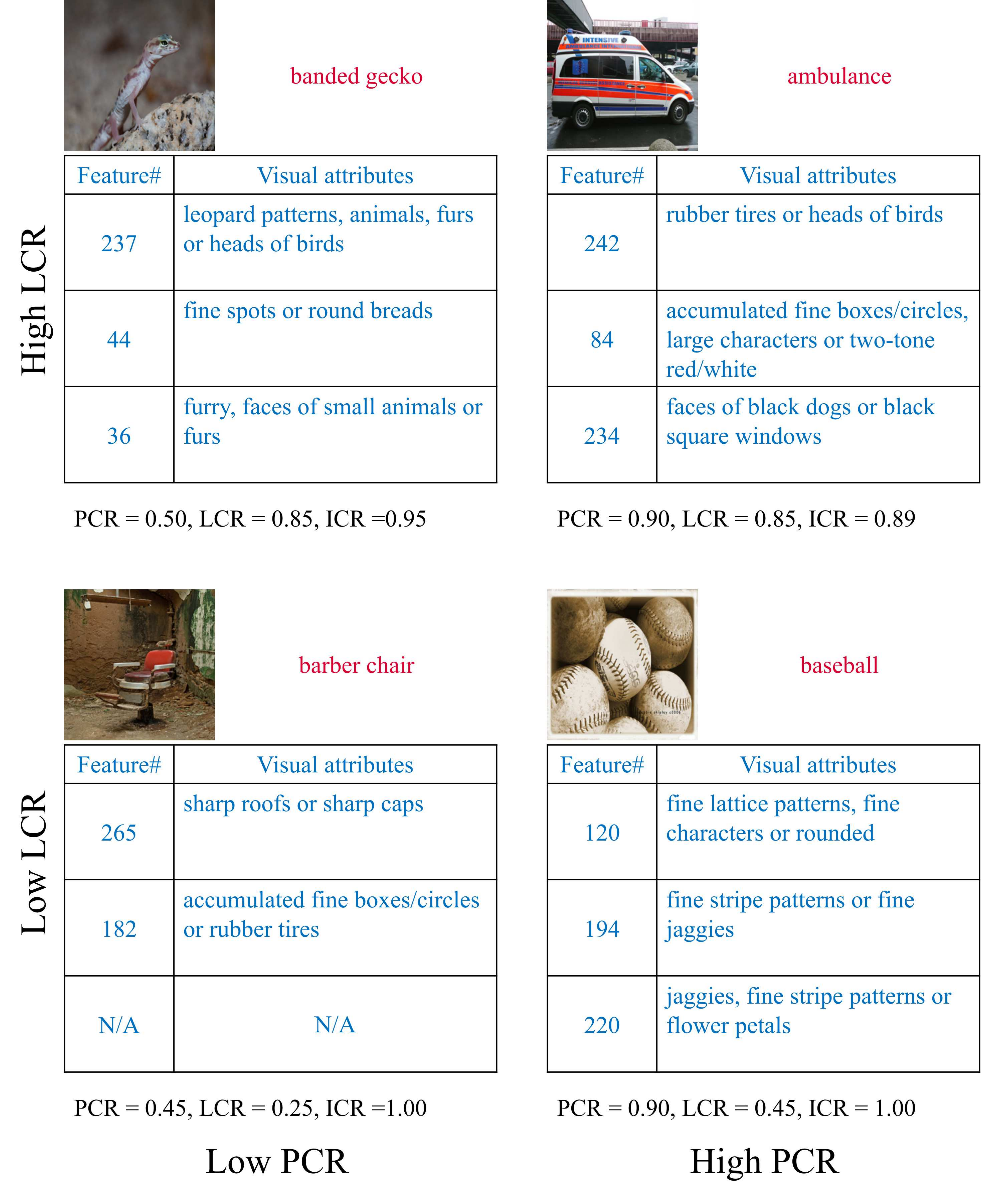} 
  \label{high_icr}
  }
  \caption{Feature and consistency analysis on images with correct inference (label). left to right: PCR increases; bottom to top: LCR increases.}
\vspace{10mm}
  \label{correctRes}
\end{figure*}

\begin{figure*}
  \centering
    \includegraphics[scale=0.1,bb=0 0 4278 2517]{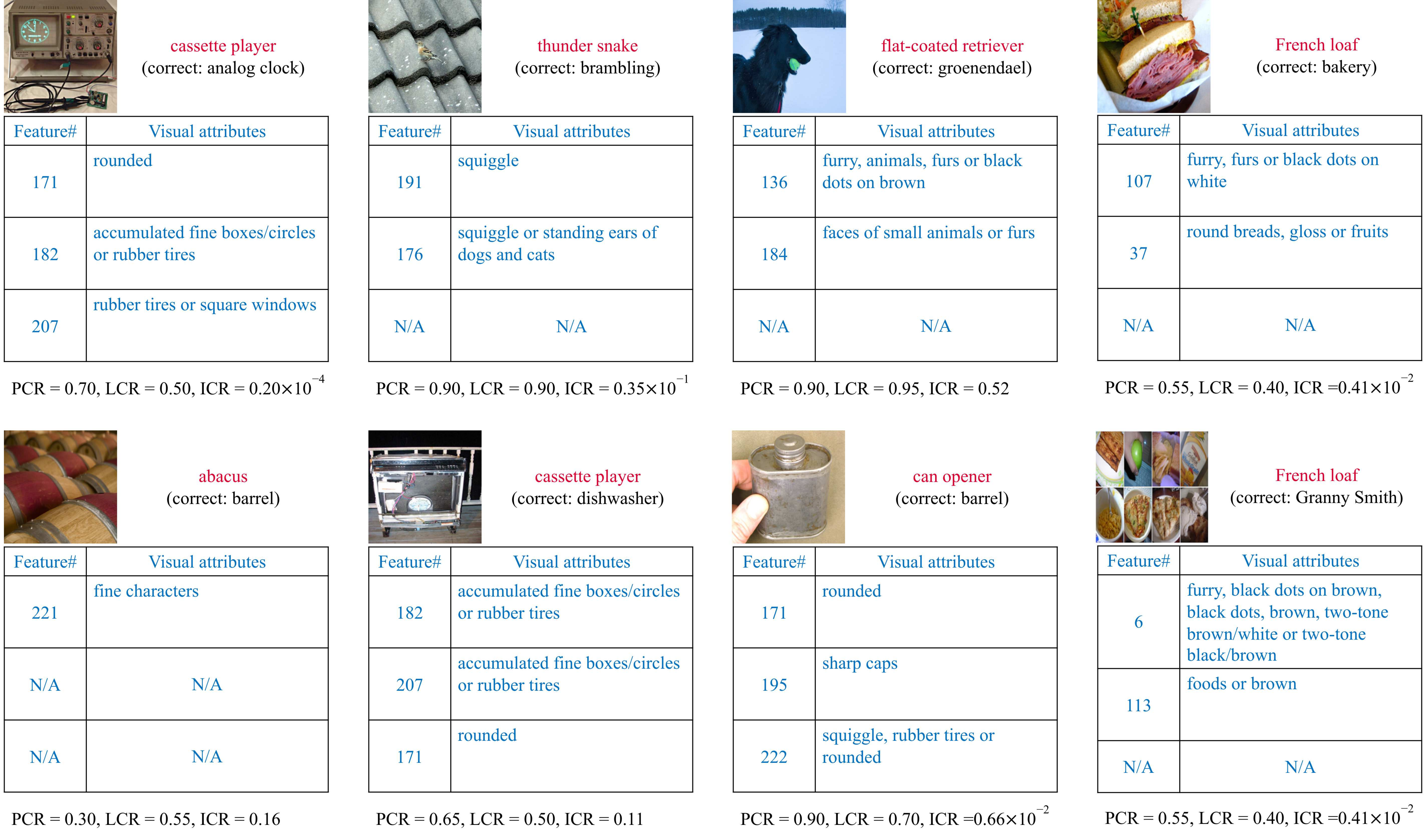} 
  \caption{Analysis results of analysis on images with incorrect inference (label)}
    \label{incorrectRes}
\end{figure*}

Let's assume that we have a CNN which we are currently train and test. 
We have inference (label) by the currently trained model, and the results of our proposed analysis. 
Figure \ref{correctRes} and Fig. \ref{incorrectRes} show results of analysis for correct inference and incorrect inference, respectively. 
Images in Fig. \ref{correctRes} and Fig. \ref{incorrectRes} are converted into $227 \times 227$ which is the actual size of the input image to CaffeNet, 
and receptive fields are omitted due to space limitation.
We walk through these results of analysis to see what we can read from them.

For the images on the left column in Fig. \ref{low_icr} which have the results of analysis in low physical consistency ratio and inference consistency ratio, the number of feature vector elements in inference features can be less than $\ell$, the maximum number of feature vector elements in the inference features.
 Human workers may have evaluated these images' physical consistency ratios low, because they saw few feature vector elements in inference features.
It is interesting that inference consistency, which is the maximum softmax probability, is also low, if the number of feature vector elements in inference features is small.
It is suggested that the inference process we hypothesized in this work is not far from actual process in neural networks.

The images on the right column in Fig. \ref{low_icr} have high physical consistency ratio and low inference consistency ratio.
The bottom one has low logical consistency ratio, because the labels of visual attributes are not appropriate.
Cassette players should have two large speakers on the left and right, and the feature maps $196$ and $171$ may represent them.
However, the labels (visual attributes) for these feature maps are rubber tires or rounded (shape), and human workers may not able to associate them.
There is room to improve the labels of visual attributes so that humans can easily comprehend the linguistic feature analysis.

The top right image in Fig. \ref{high_icr} has high physical, logical, and inference consistency. 
We see three types of visual attributes; 1) shape (fine lattice patterns, accumulated fine boxes/circles, leopard patterns), 2) color (two-tone red/white), and 
3) concrete object (black square windows, faces of small animals) in the linguistic feature analysis,
and these visual attributes are relevant to ambulance vehicles for humans, too. This is one of the the best examples.

The images on the right bottom in Fig. \ref{high_icr} has high physical and inference consistency ratios, and low logical consistency ratio.
The logical consistency ratio is low, because it is difficult for humans to associate visual attributes in the linguistic feature analysis with the inference class: baseball.
However, the inference consistency ratio, {\it i.e.} the maximum softmax probability, is $1.0$, therefore the neural network is very confident on this inference.
This example shows that the trained neural network may work with inference processes which humans cannot understand in some cases.

The images on the left top in Fig. \ref{high_icr} has high logical and inference consistency ratios, and low physical consistency ratio.
We can see fur like visual attributes in the result of linguistic feature analysis, however they are not found in the input image.
This example shows that there are features in the trained model which humans cannot understand.

The images on the left bottom in Fig. \ref{high_icr} has low physical and logical consistency ratio and high inference consistency ratio.
The linguistic feature analysis indicates sharp roofs/caps and accumulated fine boxes/circles or rubber tires,
but humans may not find these visual attributes in the image. On top of them,
even if these visual attributes are in the scene, humans cannot understand why they are associated with inference (label): barber chair.
This example shows the combination of the above two situations.
These three examples show the limitations of deep neural networks in terms of transparency.
There must be the essential complexity of deep neural network which we cannot make transparent. 

 In the second example from the left on the first row in Fig. \ref{incorrectRes}, inference (label) of CNN is snake, but the correct label is brambling, a type of bird.
 Our analysis indicates that the inference feature includes feature vector elements for "squiggle" visual attributes. 
 This is an example of understandable mistake of CNN. Although the inference (label) is incorrect, we see squiggle in the input image; and squiggle is likely to be a snake. 
 We assume that the size of the bird was too small compared to the size of squiggle patterns, and CNN may put high priority on snake class.
 If squiggle patterns, which are made by roof tiles, are larger than the bird, there is room for discussion if the ground truth class should be roof tile, rather than bird. 
 
 In the second example from the right on the first row in Fig. \ref{incorrectRes}, 
 the inference (label) of CNN is flat-coated retriever, but the correct label is groenendael. Both of them are black dogs.
 The inference features for this incorrect inference (label) produced by the CNN: flat-coated retriever are very similar to the inference features for the correct inference (label): groenendael. 
 This is another pattern of understandable mistake of CNN that the currently learnt visual attributes are not enough to distinguish between two classes.
 We need to collect training data more to acquire relevant visual attributes.
If inference (label) is incorrect with correct inference features, then it suggests insufficient training data to train relevant visual attributes for these classes. 
Possible action for this case is to collect additional training data for these classes.
If 1) inference features are correct for the input image, and 2) inference (label) is correct for the inference features, however 3) inference (label) is incorrect for the input image,
then an inaccurate ground truth label is suggested. Possible action for this case is to review and fix the ground truth label.

It is important in practice to know the actions we should take next.
Low physical consistency ratio suggests that the feature extraction part of the neural network is not well trained to capture enough visual attributes.
On the other hand, low logical consistency ratio suggests that the decision-making part of the neural network, such as classification or regression, is not well trained.
Possible action for the former case is to increase the layers in the feature extraction part, which is considered as the layers before {\tt conv5} for CaffeNet.
Possible action for the latter case is to increase the layers in decision-making part which is considered as the layers after {\tt conv5}.

\section{Conclusion}
In this paper, we developed three types of simple analysis; 1) structural feature analysis, 2) linguistic feature analysis, and 3) consistency analysis 
which improve the transparency of deep neural inference process, to address the black-box property of deep neural networks for safety critical applications.
We then evaluated and discussed our analysis methods and the results both qualitatively and quantitatively,
and introduced the usefulness of our proposed analysis by showing how to use the analysis results in the development process of deep learning models.

It is known that quantitative evaluation of the transparency of algorithms is challenging~\cite{DBLP:conf/icse/Dam0G18},
and we cannot say our work solved the problem completely.
However, deep neural inference process was black-box until now, 
and the experiments and discussion in this paper shows that our work moved it forward to transparency.
For example, there was no clue to improve a neural network when it produced incorrect inference (label).
Now our method give suggestions, or the possible next actions, such as expanding networks or collecting training data.


\bibliographystyle{spphys}       
\bibliography{ref}   


\end{document}